\PassOptionsToPackage{table}{xcolor}
\documentclass[10pt, twocolumn, letterpaper]{article}
\usepackage{cvpr_v3_1}     
\usepackage{color}
\usepackage[table]{xcolor}
\usepackage{ifthen}
\usepackage{float}
\usepackage{alltt}
\usepackage{amsmath}
\usepackage{amsthm}
\usepackage{newlfont} 
\usepackage{floatflt}
\usepackage{wrapfig}
\usepackage{fixltx2e}
\usepackage{multirow}

\usepackage{xspace}
\usepackage[export]{adjustbox}
\usepackage{mathtools, cuted}
\usepackage{CJKutf8} 
\usepackage[ruled]{algorithm2e}

\usepackage{siunitx, booktabs, lipsum}
\usepackage[flushleft]{threeparttable}

\usepackage{xcolor}
\usepackage{makecell}
\usepackage{arydshln}
\usepackage{adjustbox}
\usepackage{pifont}
\usepackage{amssymb}
\usepackage{booktabs}
\usepackage{tikz}

\usepackage{algpseudocode}
\usepackage{microtype}
\usepackage{float}
\usepackage{caption}
\usepackage{cuted}   

\newcommand{\final}{1}

\SetAlFnt{\small}
\SetAlCapFnt{\small}
\SetAlCapNameFnt{\small}
\SetAlCapHSkip{0pt}
\IncMargin{-\parindent}

\let\oldcaption\caption
\renewcommand{\caption}[2][]{\oldcaption[#1]{{\em #1} #2}}

\definecolor{SithColor}{rgb}{0.7,0,0} 
\newcommand{\qisun}[1]{{\color{SithColor} Qi: #1 $\qed$}}
\definecolor{ConsularColor}{rgb}{0,0.4,0} 
\definecolor{GuardianColor}{rgb}{0,0,0.8} 
\newcommand{\praneeth}[1]{{\color{GuardianColor} Praneeth: #1 $\qed$}}

\definecolor{zeshiBlue}{rgb}{0,0.0,0.9}

\newcommand{\yujie}[1]{{\color{ConsularColor} Yujie: #1}}
\newcommand{\warning}[1]{{\it\color{red} #1}}
\newcommand{\note}[1]{{\it\color{blue} #1}}
\newcommand{\nothing}[1]{}

\definecolor{AudioColor}{rgb}{0.56,0.34,0.62}

\definecolor{figred}{rgb}{1,0,0}
\definecolor{figgreen}{rgb}{0,0.6,0}
\definecolor{figblue}{rgb}{0,0,1}
\definecolor{figpink}{rgb}{1,0.63,0.63}

\newcommand{\new}[1]{{\color{blue} #1}}

\newcommand{\todo}[1]{{\it\color{orange} TODO: #1}}
\newcommand{\bm}[1]{\boldsymbol{#1}}

\newcommand{\bestcolor}{red!30} 
\newcommand{\secondbestcolor}{orange!30} 
\renewcommand{\circledast}{%
  \tikz[baseline=(char.base)]{%
    \node[shape=circle,draw,inner sep=1pt] (char) {\footnotesize$\ast$};}%
}

\algrenewcommand\algorithmiccomment[1]{\hfill\textcolor{gray}{$\triangleright$ #1}}
\newcommand{\shortcite}[1]{\cite{#1}}
\newcommand{\camr}[1]{{#1}}

\ifthenelse{\equal{\final}{1}}
{
\renewcommand{\qisun}[1]{}
\renewcommand{\praneeth}[1]{}
\renewcommand{\yujie}[1]{}
\renewcommand{\warning}[1]{}
\renewcommand{\note}[1]{}
\renewcommand{\todo}[1]{}

\renewcommand{\new}[1]{{#1}}

\renewcommand{\camr}[1]{#1}
}

\newcommand{\pseudocode}{Algorithm}
\floatstyle{plain}
\floatname{algorithm}{\pseudocode}

\newcommand{\filename}[1]{\url{#1}}
\newcommand{\foldername}[1]{\url{#1}}

\let\oldparagraph\paragraph
\iffalse

\renewcommand{\paragraph}[1]{\oldparagraph{\textbf{#1}.}} 
\else

\renewcommand{\paragraph}[1]{\oldparagraph{{#1}.}}
\fi

\ifdefined\email
\else
\newcommand{\email}[1]{\url{#1}}
\fi

\newcommand{\energy}{E}
\newcommand{\mass}{m}
\newcommand{\lightspeed}{c}

\newcommand{\domainsym}{\Omega}
\newcommand{\dimnumber}{d} 

\newcommand{\samplesym}{s}
\newcommand{\sample}{\samplesym{}}
\newcommand{\samplesetsym}{\mathcal{S}}
\newcommand{\sampleset}{\samplesetsym{}}
\newcommand{\sampleprime}{\samplesym^{\prime}{}}
\newcommand{\dist}{d}
\newcommand{\conflictdist}{r}
\newcommand{\union}{\oplus}
\newcommand{\failure}{f}
\newcommand{\assign}{\leftarrow}

\renewcommand{\dof}{DOF }

\graphicspath{
{images/}
{images/source_images/}
{images/1080p_wild/}
}

\usepackage{microtype}

\definecolor{cvprblue}{rgb}{0.21,0.49,0.74}
\usepackage[pagebackref,breaklinks,colorlinks,citecolor=cvprblue]{hyperref}
\usepackage{natbib}

\def\cvprPaperID{8914} 
\def\confName{CVPR}
\def\confYear{2025}

\title{DOF-GS: Adjustable Depth-of-Field 3D Gaussian Splatting for \\Post-Capture Refocusing, Defocus Rendering and Blur Removal}

\author{
    \begin{tabular}{p{3.2cm}cc}%
        Yujie Wang$^{1,2,3}$ & Praneeth Chakravarthula$^{3}$\textsuperscript{\dag} & Baoquan Chen$^{1,2}$\textsuperscript{\dag} \\
        \addlinespace
        \multicolumn{3}{c}{$^1$State Key Laboratory of General Artificial Intelligence, Peking University} \\
         \multicolumn{3}{c}{$^2$School of Intelligence Science and Technology, Peking University}\\
         \multicolumn{3}{c}{$^3$University of North Carolina at Chapel Hill}\\
    \end{tabular}
}

\begin{document}
\twocolumn[{
\renewcommand\twocolumn[1][]{#1}

\maketitle
\vspace{-3.2em}
\begin{center}
    \url{https://dof-gs.github.io/}
\end{center}

\begin{center}
    \vspace{-1.3mm}
    \setlength{\abovecaptionskip}{4pt}
    \centering
    \includegraphics[width=\textwidth]{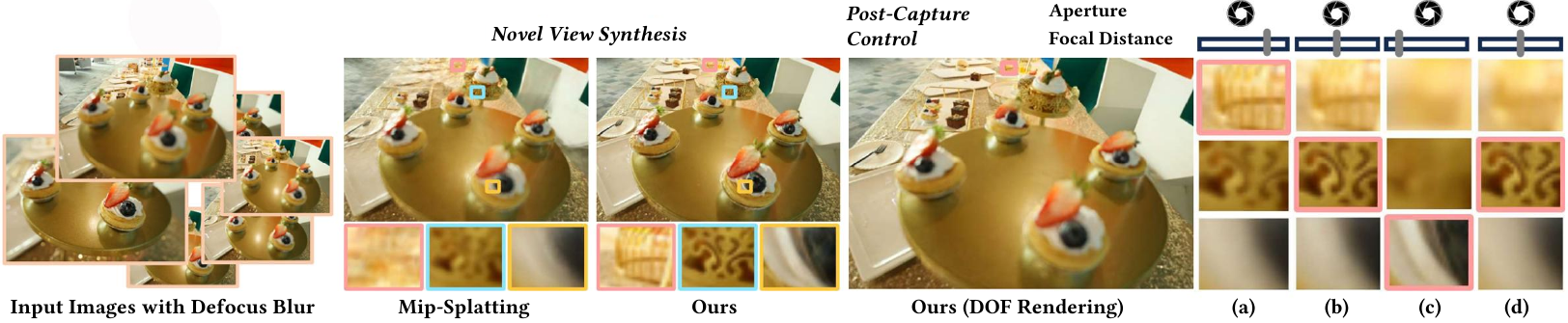}
    \captionof{figure}{
    \new{
    \camr{Our method enhances 3DGS with a finite-aperture camera model and differentiable depth-of-field (DOF) rendering, enabling post-capture DOF control after training and capturing natural defocus cues from multi-view defocused images. Differentiable DOF rendering also mitigates reconstruction artifacts caused by multi-view inconsistent defocus blur. Insets (a)-(d), with in-focus areas in pink, show images rendered under various settings: (a)-(c) share aperture but differ in focal distance; (d) matches (b)'s focal distance but has a larger aperture.}
    }}
    \label{fig:teaser}
    \vspace{2mm}
\end{center}
}]

\makeatletter
\def\@makefnmark{}
\makeatother
\footnotetext{\dag: Denotes joint advising, co-corresponding authors.}

\begin{abstract}

3D Gaussian Splatting (3DGS) techniques have recently enabled high-quality 3D scene reconstruction and real-time novel view synthesis. 
These approaches, however, are limited by the pinhole camera model and lack effective modeling of defocus effects.
Departing from this, we introduce DOF-GS --- a new 3DGS-based framework with a finite-aperture camera model and explicit, differentiable defocus rendering, enabling it to function as a post-capture control tool.
By training with multi-view images with moderate defocus blur, DOF-GS learns inherent camera characteristics and reconstructs sharp details of the underlying scene, particularly, enabling rendering of varying DOF effects through on-demand aperture and focal distance control, post-capture and optimization. 
Additionally, our framework extracts 
circle-of-confusion cues during optimization to identify in-focus regions in input views, enhancing the 
reconstructed 3D scene details. 
Experimental results demonstrate that DOF-GS supports post-capture refocusing, adjustable defocus and high-quality all-in-focus rendering, 
from multi-view images with uncalibrated defocus blur.

\end{abstract}

\section{Introduction} 


Despite the advancements made in digital photography, the need to set parameters like focus, aperture and exposure at capture-time remains a limitation,
particularly for casual photographers facing dynamic conditions.
Although methods such as light-field \cite{Ng2005plenoptic, huang2016, wei2015irregular} and multi-/coded-aperture photography \cite{Green2007, inagaki2018learning, habuchi2024time} offer some post-capture control, they typically require specialized hardware, limiting their accessibility, especially on mobile devices.
%
%
%
Recent advancements in neural 3D reconstruction techniques, such as neural radiance fields (NeRF) \cite{mildenhall2020nerf, muller2022instant, mobilenerf} and 3D Gaussian Splatting (3DGS) \cite{yang2023gs4d, duan20244d, hierarchGS} enable high-quality view synthesis from multi-view captures. 
Specifically, 
3DGS allows for realistic post-capture viewpoint control in real-time, making it attractive for a wide range of applications, including immersive, interactive experiences like augmented and virtual reality (AR/VR).
Beyond viewpoint control, however, adjusting parameters like focus and aperture post capture, and rendering their associated defocus and depth-of-field (DOF) effects, remain challenging in these approaches due to their inherent pinhole camera model assumption.


Some recent variants of these methods although model defocus blur, they are insufficient to support post-capture control of focal and aperture parameters due to their black-box modeling.
For instance, previous NeRF-based techniques  \cite{ma2022deblur, lee2023dp, peng2023pdrf} simulate defocus blur by estimating per-ray blur kernels and aggregating neighboring rays. However, these methods do not explicitly model a finite aperture camera, overlooking the relationships between point spread functions across pixels, limiting controllable defocus rendering, post-optimization.
Moreover, these approaches are computationally expensive, with training durations ranging from hours to nearly a day \cite{peng2023pdrf}, and slow novel view synthesis with speeds typically falling below $1$ FPS even at moderate resolutions like $600 \times 400$ pixels \cite{lee2024deblurring}. 
Recently, 3DGS-based methods \cite{lee2024deblurring, peng2024bags} have attempted to model defocus blur with neural networks that predict per-Gaussian shape offsets or per-pixel blur kernels in image space. 
%
However, these techniques still lack built-in support for intuitive post-capture 
control through adjustable camera parameters.

In this paper, we introduce DOF-GS, a 3DGS-based framework that \textit{incorporates a finite-aperture camera model} for modeling and rendering depth-of-field effects in reconstructed scenes, 
moving beyond the pinhole camera limitations in existing methods.
By introducing a finite-aperture camera model, DOF-GS enables the
adjustment of focal distance and aperture to render defocus effects, correctly representing depth-related blurring via the circle-of-confusion (CoC). 
Inspired by the surface-splatting based technique \cite{Krivanek2003}, our framework efficiently renders defocus effects by blurring individual Gaussians based on the learnable CoC cues, before compositing the final image via rasterization. Since Gaussian points optimized on sharp images do not naturally form defocus effects under this rendering scheme,
we guide the model using defocus blur from captured images to enable smooth and realistic defocus rendering. 
To support optimization of
Gaussian points and \textit{learning the camera characteristics 
with defocused images}, we design the DOF rendering process to be \textit{differentiable}. Additionally, to counteract reduced sharpness due to defocus blur in multi-view inputs, we enhance scene details by leveraging CoC cues emerging during optimization. 

To summarize, our approach extends the 3DGS framework with three key enhancements:
1) \textbf{CoC-guided DOF rendering:} A novel rendering process integrated into the 3DGS rasterizer, allowing efficient and differentiable modeling of defocus effects. 
2) \textbf{Learnable camera parameters:} Two adaptive parameters per view to dynamically fit defocus levels in training views and enable flexible DOF adjustments, post-optimization. 
3) \textbf{CoC-guided detail enhancement:} A lightweight CoC-guided neural network that predicts in-focus mask for each view, which are utilized to enhance details within All-in-Focus renderings during optimization.

We experimentally test our framework on scenes from the multi-view defocus blur dataset \cite{ma2022deblur}. Results demonstrate that our method effectively utilizes defocus blur from multi-view input defocused images, enabling on-demand DOF rendering and post-capture refocusing (Figure \ref{fig:teaser}). 


\section{Related Work}

\noindent \textbf{Neural Radiance Field and 3D Gaussian Splatting.}
By bridging 2D multi-view images with the underlying 3D scene via differentiable volume rendering, NeRF \cite{mildenhall2020nerf} has made great progress in 3D reconstruction and novel view synthesis. In recent years, NeRF has been extended to handle dynamic scenes \cite{pumarola2020d, liu2022devrf} and enhanced with explicit representations for improved performance \cite{muller2022instant, mobilenerf, plenoxels, kulhanek2023tetra}. 
Recently, Kerbl et al. \cite{3dgs} introduce the 3D Gaussian Splatting technique, achieving high-quality real-time renderings at notably high resolutions. 
Subsequent developments have further enhanced its capabilities by handling anti-aliasing \cite{mip-splatting}, dynamic scenes \cite{yang2023gs4d, duan20244d}, and improving rendering speeds for large-scale environments \cite{hierarchGS}. However, most NeRF- and 3DGS-based approaches prioritize producing sharp images, often overlooking modeling and rendering of defocus blur.

\noindent \textbf{Defocus Deblurring in NeRF and 3DGS.}  
Some studies simulate defocus blur during NeRF or 3DGS training to address multi-view inconsistencies and blurry reconstructions caused by defocus blur in multi-view inputs. For example,  
NeRF-based methods \cite{ma2022deblur, lee2023dp, peng2023pdrf} simulate defocus by predicting pixel-wise blur kernels per view and aggregating multiple rays at each pixel.
However, these methods lack an explicit finite-aperture camera model, leading to flexible but arbitrarily predicted blur kernels without accounting for relationships among point spread functions across different pixels, thus limiting post-capture focus and aperture control. 
Moreover, the aggregation of multiple rays in volume rendering demands extensive sampling, making these methods highly computationally expensive.
Recently, some 3DGS methods \cite{lee2024deblurring, peng2024bags} integrate defocus modeling to mitigate the effect of defocus blur and improve rendering efficiency. 
These methods use neural networks to predict per-Gaussian shape offsets or dense pixel-wise kernels in image space to fit the defocus observed in training views.
However, due to this black-box modeling, they still lack inherent support for post-capture focus and aperture control.


\noindent \textbf{Defocus Rendering in NeRF.}
Defocus blur is typically rendered by simulating image formation with a finite-aperture camera model.
Existing methods like RawNeRF \cite{rawnerf} and NeRFocus \cite{nerfocus} synthesize DOF effects using NeRF, but assume all-in-focus (AiF) inputs. 
The closest NeRF-based counterpart to our approach, DoF-NeRF \cite{dof-nerf}, also aims at controllable DOF rendering post-training but suffers from high computational costs of volume rendering, and differs from our method in defocus blur modeling and simulation. 
To address the limitations of the pinhole camera model in existing 3DGS methods, we extend the 3DGS framework to include a finite-aperture model with two learnable parameters: aperture size and focal distance.
Our approach enables defocus rendering by applying defocus effects to individual Gaussians before compositing the final image, inspired by surface-splatting techniques \cite{Krivanek2003}. 
Notably different from \cite{Krivanek2003}, our rendering process is fully differentiable, enabling optimization of associated camera parameters---such as aperture diameter and focal distance---alongside the underlying 3D scene based on multi-view defocused images.

\begin{figure*}[ht!]
\begin{center}
\includegraphics[width=.95\linewidth]{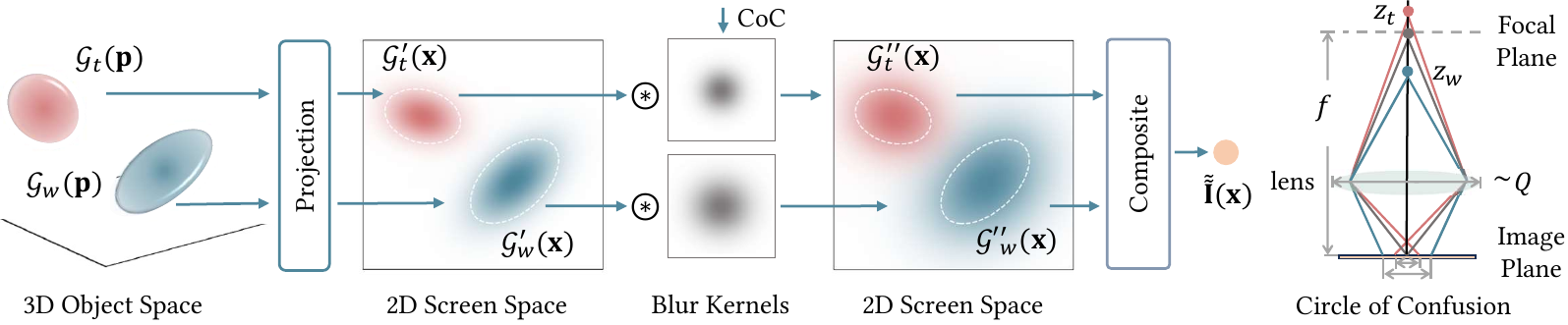}
\end{center}
\vspace{-3mm}
\caption{
\camr{Illustration of the depth-of-field rendering and camera model. During the DOF rendering (left), 3D Gaussians $\mathcal{G}_k$ are projected to 2D screen space. Each projected 2D Gaussian $\mathcal{G}^{\prime}_k$ is then convolved with a blur kernel, and the final color is composited from convolved Gaussians $\mathcal{G}^{\prime\prime}_k$. White dashed lines highlight convolution effects. The blur kernel for each 2D Gaussian is determined by the radius of its Circle-of-Confusion (CoC), which results from the finite-aperture camera model (right). With the adopted camera model with an aperture parameter $Q$, which is not a pinhole, object points deviating from the focal distance $f$ form a region, known as the CoC, rather than a point.}
}
\vspace{-1mm}
\label{fig:dof_gaussian}
\end{figure*}

\section{Method} \label{sec:approach}
Our approach, illustrated in \Cref{fig:method}, 
\new{reconstructs a 3D scene and enables post-capture depth-of-field (DOF) rendering from}
multi-view images $\{{\mathbf{I}_m}\}_{m=1}^{M}$ exhibiting varying and arbitrary defocus blur.
Building on the 3DGS framework \cite{3dgs}, we represent the scene with a set of anisotropic 3D Gaussian points $\{\mathcal{G}_k\}_{k=1}^{K}$, optimize camera parameters to allow flexible, post-capture adjustments, while also mitigating defocus blur effects for fine 3D scene reconstruction.
In \Cref{sec:3dgs}, we outline the scene representation, 3DGS rasterization-based rendering process, and our camera and physical models for depth-of-field. 
\Cref{sec:diff_dof} details our differentiable DOF rendering, crucial for simulating defocus blur in training views. 
Finally, \Cref{sec:joint_opt} describes an optimization strategy to enhance scene detail in reconstruction \new{by using defocus cues derived from the DOF rendering}.

\subsection{Preliminaries}\label{sec:3dgs}

\subsubsection{3D Gaussian Splatting}
The 3DGS technique \cite{3dgs} is a point-based method where 3D scenes are explicitly represented through an array of 3D Gaussian points given by:
\vspace{-2mm}
\begin{equation}
     \mathcal{G}_k(\mathbf{p})=\text{exp}\left(-\frac{1}{2}\left(\mathbf{p}-\mathbf{u}_k\right)^{T}\bm{\Sigma}_k^{-1}\left(\mathbf{p}-\mathbf{u}_k\right)\right),
\end{equation}
where $\mathbf{p}$ denotes 3D points in world space.
Specifically, each Gaussian is defined by following attributes: a center position $\mathbf{u}_k$, covariance matrix $\bm{\Sigma}_k$ derived from anisotropic scaling $\mathbf{s}_k$ and a quaternion vector $\mathbf{q}_k$, as well as opacity $o_k$ and spherical harmonics coefficients $\mathbf{h}_k$. Ultimately, a scene is parameterized as $\mathcal{S}=\{\mathcal{G}_k: \mathbf{u}_k, \mathbf{q}_k, \mathbf{s}_k, o_k, \mathbf{h}_k\}_{k=1}^{K}$. This method realizes high-quality efficient rendering through a customized Gaussian rasterization process. Following the approach by Zwicker et al. \shortcite{zwicker2001ewa}, the 3D Gaussian points are first projected to 2D screen space, during which the Jacobian $\mathbf{J}$ of the 
projective transformation is calculated, and the covariance matrix $\bm{\Sigma}_k^{\prime}$ of the 2D Gaussian after projection is computed as 
\begin{equation}
\Sigma_k^{\prime} = \mathbf{J}\mathbf{W}\bm{\Sigma}_k \mathbf{W}^T \mathbf{J}^T.
\end{equation}
$\mathbf{W}$ represents the view matrix transforming points from world space to camera space. The 2D Gaussian in screen space is formulated as:
\vspace{-2mm}
\begin{equation}
    \mathcal{G}_k^{\prime}(\mathbf{x})=\text{exp}\left(-\frac{1}{2}\left(\mathbf{x}-\mathbf{u}_k^{\prime}\right)^{T}\left(\bm{\Sigma}_k^{\prime}\right)^{-1}\left(\mathbf{x}-\mathbf{u}_k^{\prime}\right)\right),
\end{equation}
where $\mathbf{u}_k^{\prime}$ represents the 2D center position, post-projection.
Although $\mathcal{G}_k^{\prime}$ has an infinite support in theory, it is evaluated on a limited range with a cutoff radius $t$ in practice.
Therefore, each pixel $\mathbf{x}$ is associated with only a small number $\mathcal{N}_x$ of the Gaussians in the scene.
Finally, the projected Gaussians are rendered with alpha blending:
\vspace{-2mm}
\begin{equation}\label{eq:blending}
    \tilde{\mathbf{I}}(\mathbf{x}) = \sum_{i=1}^{\mathcal{N}_x} T_i \alpha_i {\mathbf{c}}_i, \:\:\:\: \alpha_i = {\mathcal{G}}^{\prime}_i(\mathbf{x}),
\end{equation}
where $T_i = \prod_{j=1}^{i-1}(1-\alpha_j)$ is transmittance, ${\mathbf{c}}_i$ denotes the view-dependent color of $i$-th Gaussian associated with the queried pixel $\mathbf{x}$, \new{$\tilde{\mathbf{I}}(\mathbf{x})$ is rendered color on pixel $\mathbf{x}$.} As \Cref{eq:blending} is fully differentiable, 3DGS reconstructs a 3D scene by 
\new{comparing rendered views against training views:}
\vspace{-2mm}
\begin{equation}\label{eq:loss_ori}
\min_{\mathcal{S}}\sum_{m=1}^{M}\mathcal{L}_{\text{rec}}(\tilde{\mathbf{I}}_m, \mathbf{I}_m),
\end{equation}
\new{where $\mathcal{L}_{\text{rec}}$ measures errors between two images through a weighted combination of MSE and DSSIM loss: $\mathcal{L}_{\text{rec}}(\mathbf{I}_a, \mathbf{I}_b)=||\mathbf{I}_a-\mathbf{I}_b||+\lambda_{\text{dssim}}\mathcal{L}_{\text{dssim}}(\mathbf{I}_a, \mathbf{I}_b)$ and $m$ is the index running over the training views.}

\subsubsection{Camera model for DOF rendering} \label{sec:cam_model}
We use a simplified
thin lens model for rendering DOF. A camera model is typically specified by the following attributes: focal length of the lens $F$, aperture diameter $A$, and focal distance of the object $f$. As illustrated in \Cref{fig:dof_gaussian}, the object points deviating from the focal plane appear out-of-focus, and forms the circle-of-confusion (CoC).
The radius of CoC for each point is determined by the depth $z_o$ of the object point and the camera attributes aforementioned, and is given by
\vspace{-3mm}
\begin{equation}\label{eq:coc}
    R^{(\text{coc})}=\frac{1}{2}\:FA\frac{|z_o-f|}{z_o(f-F)}.
\end{equation}
As the lens focal length $F$ is often much smaller than the focal distance $f$ and depths of object points, we simplify \Cref{eq:coc} to
\vspace{-3mm}
\begin{equation}\label{eq:coc_final}
    R^{(\text{coc})} = \frac{1}{2} \:  Q  \: |\frac{1}{z_o} - \frac{1}{f}|,
\end{equation}
where $Q=FA$ is a simplified aperture parameter.

\subsection{Differentiable DOF Rendering} \label{sec:diff_dof}
\new{To effectively leverage the guidance of natural defocus blur present in inputs, while also mitigating possible reconstruction artifacts caused by defocus blur (as shown in \Cref{fig:teaser}), we develop a differentiable DOF rendering pipeline.}
Specifically, we assume that a blurred image can be generated by first applying depth-dependent blur to each Gaussian in the scene, followed by rendering the image from these blurred Gaussians \cite{Krivanek2003}. 
We illustrate the rendering process in \Cref{fig:dof_gaussian} and detail it below.

As illustrated in \Cref{fig:dof_gaussian}, each 2D Gaussian $\mathcal{G}_k^{\prime}$, representing the 2D projection of a 3D Gaussian point, is convolved with a Gaussian blur kernel proportional to the CoC determined by the camera model described in \Cref{sec:cam_model}. We assume uniform depth across each 2D Gaussian's support, which is set as $z_k$, \emph{i.e.}, the z-coordinate of the transformed center position in camera space. With the CoC radius $R^{(\text{coc})}_{k}$ calculated from \Cref{eq:coc}, we construct the blur kernel as:
\begin{equation}
    \mathcal{G}^{\text{(coc)}}_{k} = \frac{1}{2}\exp{\left(\mathbf{x}^{T}\bm{\Sigma}_{k}^{\text{(coc)}}\mathbf{x}\right)}, \:\:\:\bm{\Sigma}_{k}^{\text{(coc)}} = a\mathbf{I}=\begin{bmatrix} a & 0 \\ 0 & a \end{bmatrix}.
\end{equation}
$a$ is a scalar, $\mathbf{I}$ is an identity matrix, and $\boldsymbol{\Sigma}_{k}^{\text{(coc)}} = a\mathbf{I}$ means that $\mathcal{G}^{\text{(coc)}}_{k}$ is isotropic. \new{We determine $a$'s value by solving}
\begin{equation}
    \min_{a}\Bigg|\Bigg|\frac{1}{\pi \left(R_{k}^{(\text{coc})}\right)^2} - \frac{1}{2 \pi a} \exp{\left(-\frac{1}{2}\frac{\mathbf{x}^{T}\mathbf{x}}{a}\right)}\Bigg |\Bigg|_2,
\end{equation}
which gives $a=\frac{1}{2ln4}\left(R_{k}^{(\text{coc})}\right)^2$, making $\mathcal{G}_{k}^{(\text{coc})}$ approximate a uniform intensity distribution within the CoC.
Applying Gaussian kernels for blur convolution maintains the integrity of color composition, as convolving two Gaussians results in another Gaussian, preserving consistency with the original rasterization process.
The convolved 2D Gaussian is defined by $\mathcal{G}^{\prime\prime}_k = \mathcal{G}^{\prime}_k  \: \circledast  \: \mathcal{G}^{\text{(coc)}}_{k}$, where $\circledast$ denotes convolution.
\Cref{fig:dof_gaussian} illustrates the changes introduced by the convolution to individual 2D Gaussians. Since the convolution changes the associations between the 2D Gaussians and pixels, we reevaluate the number of Gaussians linked to each pixel, denoted as $\mathcal{N}_x^{\prime}$. Subsequently, the color of each pixel $\mathbf{x}$ is obtained by accumulating contributions from these associated Gaussians. This process produces the output image ${\tilde{\tilde{\mathbf{I}}}}$, resulting in a refined optimization objective for scene reconstruction:
\vspace{-2mm}
\begin{equation}\label{eq:obj_final}
    \min_{
    \{\mathcal{S},\:\{f_m,\: Q_m\}_{m=1}^{M}\}}\:\sum_{m=1}^{M}\mathcal{L}_{\text{rec}}(\Tilde{{\tilde{\mathbf{I}}}}_m, {\mathbf{I}}_m).
\end{equation}
As indicated in \Cref{eq:obj_final}, the optimizable parameters now include both the underlying 3D scene $\mathcal{S}$ and the camera parameters $\{f_m, Q_m\}_{m=1}^{M}$ for training views. To ensure the efficiency and differentiability, we incorporate Gaussian convolution and derivative calculations into the CUDA-customized rasterization process \cite{3dgs}. Further details on the derivations and the initialization strategy for the camera parameters can be found in the Supplementary Material.

\begin{figure*}[htbp!]
\begin{center}
\includegraphics[width=.99\linewidth]{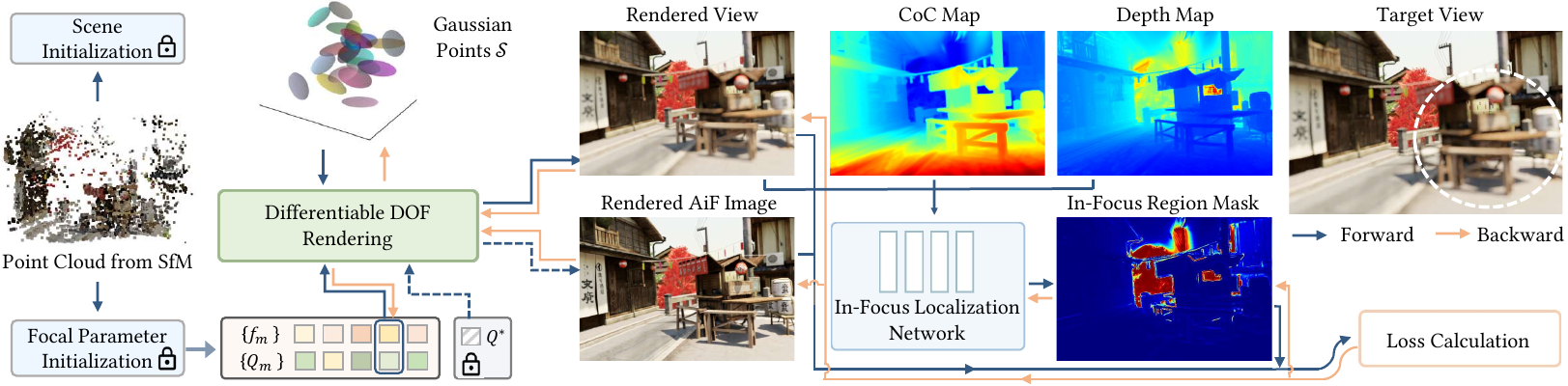}
\end{center}
\vspace{-2mm}
\caption{
Pipeline of the proposed DOF-GS. We start with camera poses and an initial sparse point cloud estimated from defocused images. For each view, we introduce and initialize two learnable parameters: focal distance and aperture parameter. \camr{During optimization, for each sampled view $m$, we render a defocused image utilizing the DOF rendering scheme with camera parameters $\{f_m, Q_m\}$ to fit the target view. Meanwhile, we render an All-in-Focus (AiF) image with the fixed aperture parameter $Q^{*}$. To enhance scene details by appropriate AiF image supervision, we introduce an In-Focus Localization Network that utilizes the rendered CoC map and other cues to localize the in-focus regions within the target view. The underlying 3D scene, camera parameters, and network parameters are updated via backpropagation.}
}
\label{fig:method}
\end{figure*}
\begin{figure}
\begin{center}
\includegraphics[width=\linewidth]{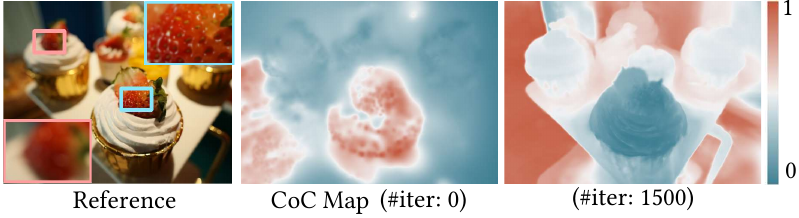}
\end{center}
\vspace{-4mm}
\caption{The rendered Circle-of-Confusion map quickly becomes correlated with pixel-wise blur levels in the reference view.
}
\vspace{-4mm}

\label{fig:fcoc_vis}
\end{figure}

\noindent \textbf{Depth and CoC map rendering. } As illustrated in \Cref{fig:method}, the differentiable DOF rendering not only renders the color image ${\tilde{\tilde{\mathbf{I}}}}$ but also a depth map, expressed as:
\vspace{-2mm}
\begin{equation}
\small
    {\tilde{\tilde{\mathbf{D}}}}(\mathbf{x}) = \sum_{i=1}^{\mathcal{N}_x^{\prime}}, T^{\prime}_i \alpha_i^{\prime} z_i, \:\: \alpha_i^{\prime} = \mathcal{G}^{\prime\prime}_i(\mathbf{x}),\:\: T^{\prime}_i=\prod_{j=1}^{i-1}(1-\alpha_j^{\prime}),
\end{equation}
where $z_i$ is the z-coordinate of the transformed center position in camera space. Moreover, we also render a CoC map $\mathcal{M}^{(\text{coc})}$ as:
\vspace{-2mm}
\begin{equation}
\small
    \mathcal{M}^{(\text{coc})}(\mathbf{x}) = \sum_{i=1}^{\mathcal{N}_x^{\prime}} T^{\prime}_i \alpha_i^{\prime} R_{i}^{(\text{coc})}, \:\: \alpha_i^{\prime} = \mathcal{G}^{\prime\prime}_i(\mathbf{x}),\:\: T^{\prime}_i=\prod_{j=1}^{i-1}(1-\alpha_j^{\prime}).  
\end{equation}
Empirical observations indicate that the rendered CoC map $\mathcal{M}^{(\text{coc})}$ becomes
highly correlated with the blur levels of pixels in training image as the optimization progresses, as shown in  \Cref{fig:fcoc_vis}.

\subsection{CoC-Guided Detail Enhancement} \label{sec:joint_opt}
\new{
The presence of large out-of-focus areas in training images can lead to the reconstructed scene prioritizing defocus blur effects, potentially compromising the scene detail clarity.
To address this, we introduce a CoC-guided detail enhancement strategy that strengthens detail recovery by also supervising All-in-Focus images rendered by setting aperture to $0$.
Since explicit labels for in-focus regions are unavailable, we devise a lightweight neural module to estimate in-focus regions with the guidance of rendered CoC maps, which correlate with pixel-blurriness in training views. The neural module is called In-Focus Localization Network (ILN).
Subsequently, we leverage in-focus masks output from ILN to supervise rendered All-in-Focus images.}


\noindent \textbf{In-Focus Localization Network. } 
The In-Focus Localization Network is designed to guide the optimization to ensure proper focus on in-focus regions. 
\new{
We use rendered CoC map $\mathcal{M}_m^{\text{(coc)}}$ as the primary input for the ILN to leverage the cues for per-pixel blur extent.
We further augment the input with the rendered defocused image $\tilde{\tilde{\mathbf{I}}}_m$ and depth map $\tilde{\tilde{\mathbf{D}}}_m$ to provide contextual color and depth cues.}
Additionally, the training view index and pixel coordinates are also fed into the ILN through positional encoding, to assist in spatial localization.
To minimize the computational overhead, we design the ILN to be lightweight, consisting of four 2D convolutional layers with a Sigmoid activation function. 
The ILN outputs a mask $\mathcal{M}_m^{*}$ that indicates the in-focus regions, which is given by
\vspace{-2mm}
\begin{equation}
\mathcal{M}_m^{*}=\text{ILN}\left({\tilde{\tilde{\mathbf{I}}}}_m,{\tilde{\tilde{\mathbf{D}}}}_m,\:\mathcal{M}_m^{(\text{coc})}, \text{PE}(m), \text{PE}(\textbf{x})\right),
\end{equation}
where $\text{PE}(\cdot)$ denotes positional encoding. 
As the ground truth masks for in-focus regions are not available, we regularize $\mathcal{M}_m^{*}$ using CoC map by a relaxed relative loss:
\vspace{-2mm}
\begin{equation}
\small
    \mathcal{L}_{\text{mk}}=1 - \frac{\text{Cov}(1-\mathcal{M}_m^{*}, \mathcal{M}_m^{(\text{coc})})}{\sqrt{\text{Var}(1-\mathcal{M}_m^{*})\text{Var}(\mathcal{M}_m^{(\text{coc})})}},
\end{equation}
where the right term represents Pearson correlation that measures the distributional difference between the estimated in-focus region mask and the rendered CoC map, while disregarding scale inconsistencies.
The loss $\mathcal{L}_{\text{mk}}$ guides $\mathcal{M}_m^{*}$ to have higher values for in-focus regions and lower values for out-of-focus regions.

\vspace{-1mm}
\setlength{\tabcolsep}{1pt}
\renewcommand{\arraystretch}{0.8}
\begin{figure*}[t]
\begin{center}
\begin{tabular}{ccccccc}
\includegraphics[height=1.72cm]{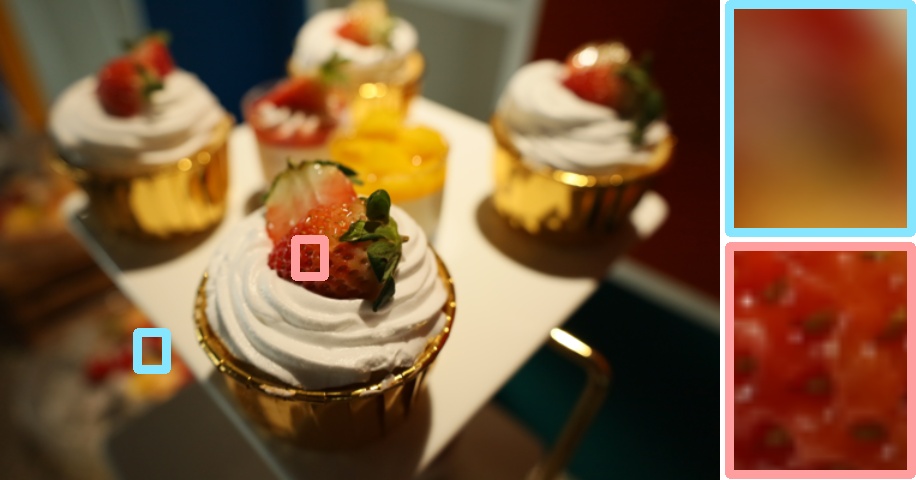}
&\includegraphics[height=1.72cm]{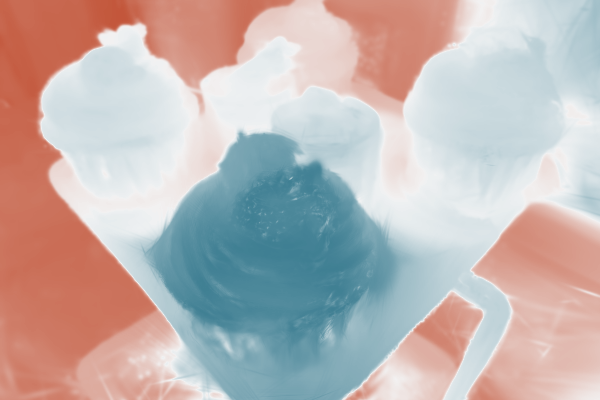}
&\includegraphics[height=1.72cm]{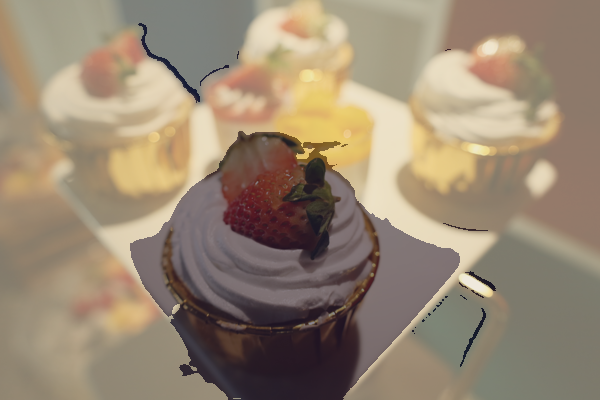}
&\includegraphics[height=1.72cm]{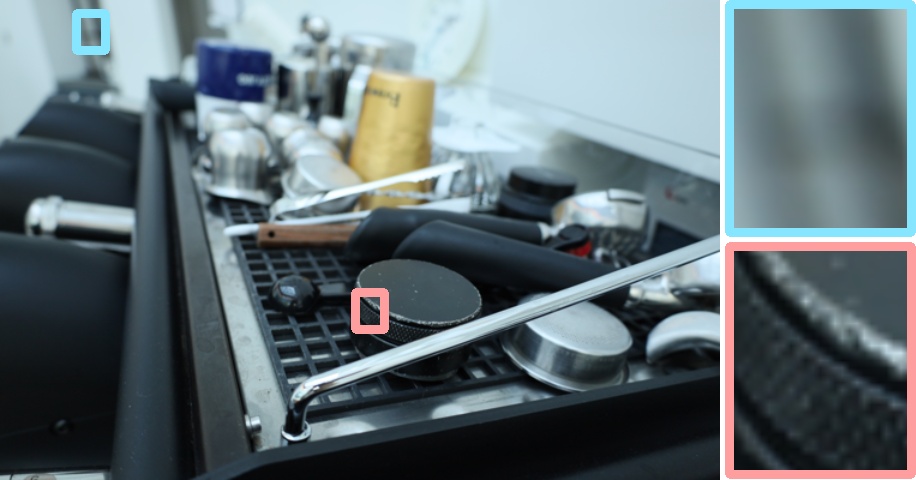}
&\includegraphics[height=1.72cm]{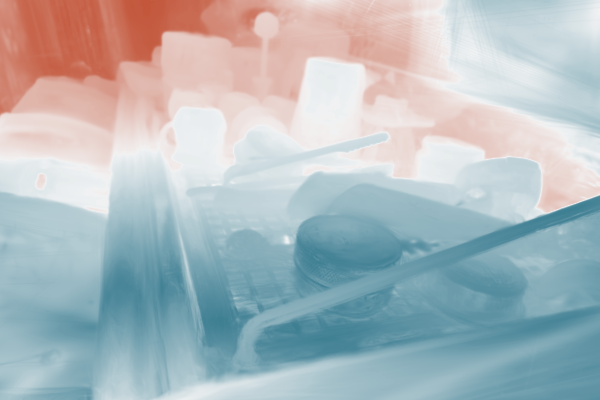}
&\includegraphics[height=1.72cm]{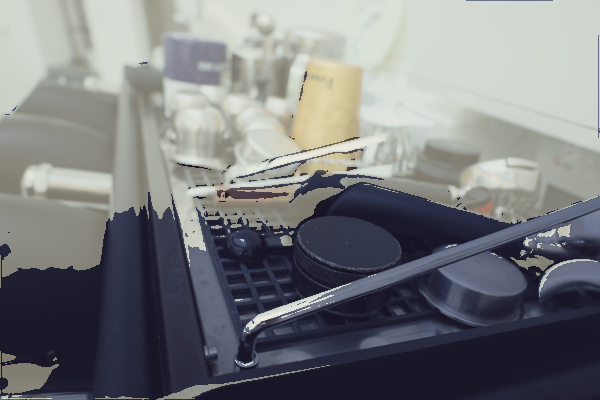}
&\includegraphics[height=1.72cm]{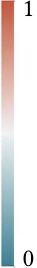}
\\
\includegraphics[height=1.72cm]{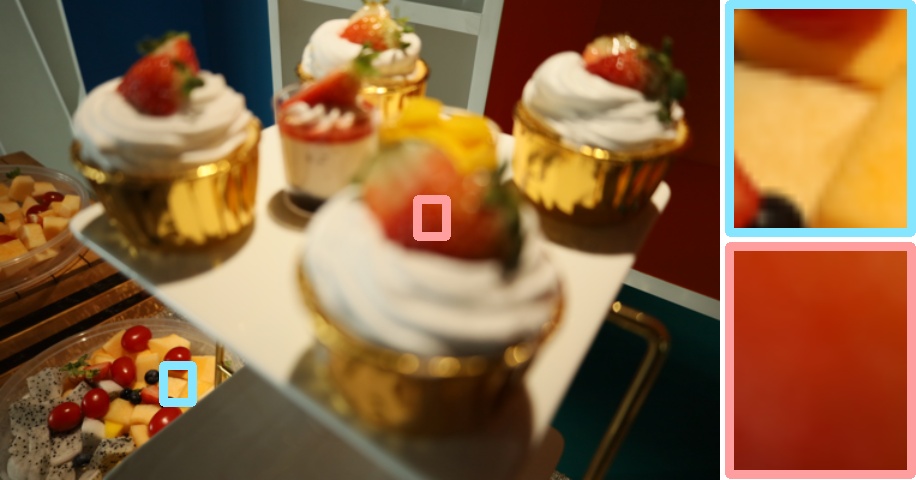}
&\includegraphics[height=1.72cm]{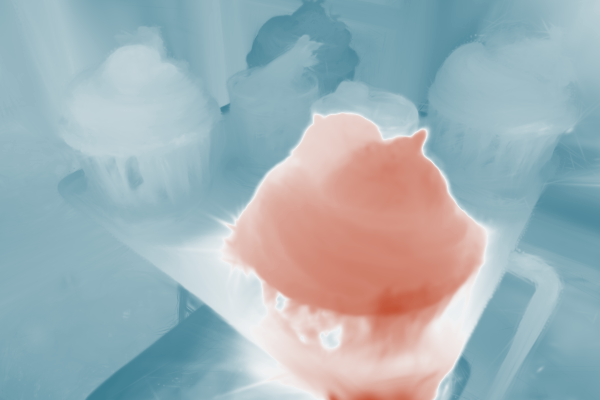}
&\includegraphics[height=1.72cm]{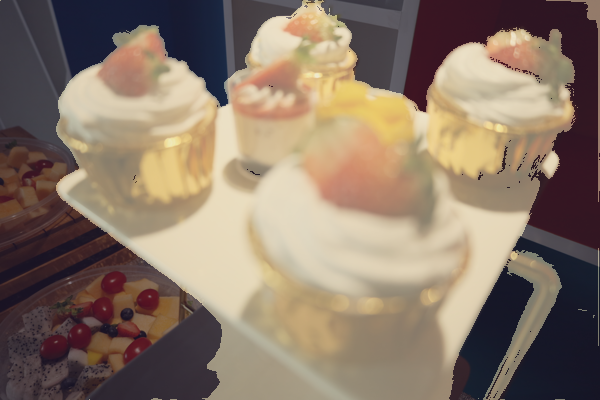}
&\includegraphics[height=1.72cm]{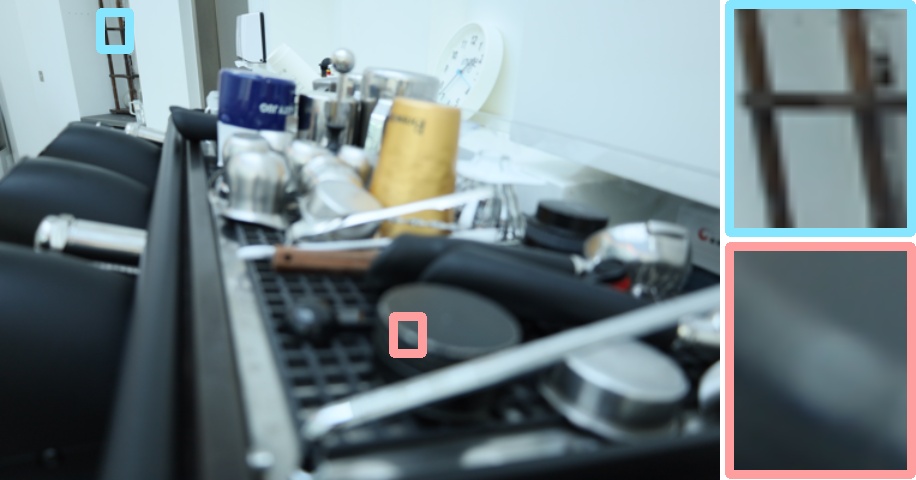}
&\includegraphics[height=1.72cm]{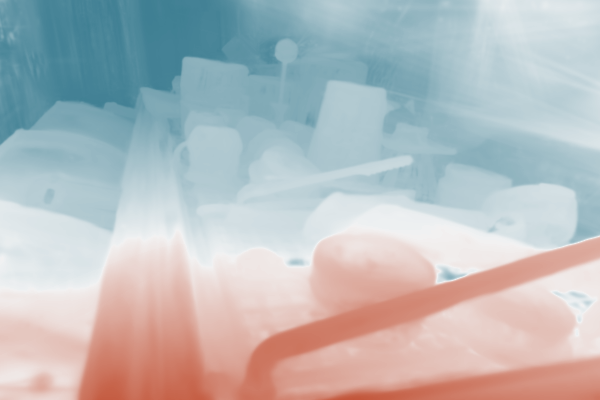}
&\includegraphics[height=1.72cm]{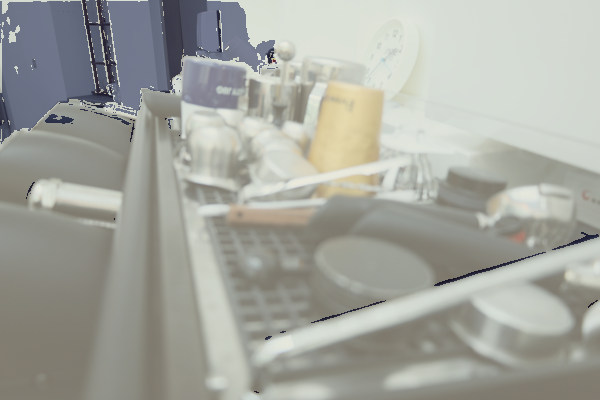}
&\includegraphics[height=1.72cm]{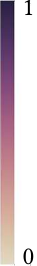}
\\
\small{Training View} & \small{CoC Map} & \small{In-Focus Mask} & \small{Training View} & \small{CoC Map} & \small{In-Focus Mask} & \\  
\vspace{-8mm}
\end{tabular}
\end{center}
\caption{
The CoC maps correlate with the depth-of-field (DOF) effects in the training views, while the predicted in-focus regions effectively locate the in-focus areas, blended with the reference image for better visualization. The CoC maps are normalized to the range $[0, 1]$.
}
\label{fig:coc_mask}
\vspace{-2mm}
\end{figure*}
\vspace{1mm}
\noindent\textbf{Optimization Objectives. }
In each iteration, we render both a defocused image $\tilde{\tilde{\mathbf{I}}}_m$ and an All-in-Focus image $\tilde{\tilde{\mathbf{I}}}_m^{*}$ from the sampled viewpoint. 
The defocused image
is rendered with the learnable aperture parameter $Q_m$ and focal distance $f_m$, while the All-in-Focus image is rendered with a fixed aperture parameter $Q^{*}=0$. With the estimated in-focus mask $\mathcal{M}_m^{*}$, 
we composite an intermediate image $\tilde{\tilde{\mathbf{I}}}_m^{c}$ by blending in-focus regions from $\tilde{\tilde{\mathbf{I}}}_m^{*}$ and out-of-focus regions from $\tilde{\tilde{\mathbf{I}}}_m$:
\vspace{-2mm}
\begin{equation}
\tilde{\tilde{\mathbf{I}}}_m^{c} = \mathcal{M}_m^{*} \tilde{\tilde{\mathbf{I}}}_m^{*} + (1-\mathcal{M}_m^{*}) \tilde{\tilde{\mathbf{I}}}_m.
\end{equation}
Then we calculate a loss $\mathcal{L}_{\text{detail}}$ by comparing the composited image against the training image:
\vspace{-2mm}
\begin{equation}
\mathcal{L}_{\text{detail}} = ||\tilde{\tilde{\mathbf{I}}}_m^{c} - {\mathbf{I}}_m|| + \lambda_{\text{dssim}}\mathcal{L}_{\text{dssim}}(\tilde{\tilde{\mathbf{I}}}_m^{c}, {\mathbf{I}}_m).
\end{equation}
\new{This loss enables us to supervise All-in-Focus renderings with in-focused regions from input images. It allows optimizing masks $\mathcal{M}_m^{*}$ with gradients weighted by errors from both all-in-focus and defocused renderings, yielding smaller values for blurry pixels in input images. Meanwhile, this strategy is also beneficial in ensuring objects positions, especially in the transition regions, are correctly represented by achieving smooth transitions between the regions extracted from the rendered defocused image and the All-in-Focus image. }
Furthermore, as the errors in the rendered images decrease during the optimization process, we encourage the in-focus mask to become more binary by incorporating a regularizer:
\begin{equation}\label{eq:m-reg}
\vspace{-1.5mm}
 L_{\text{reg}} = -\mathcal{M}_m^{*}\log\left(\mathcal{M}_m^{*}\right).
 \end{equation}
This regularizer encourages ILN to produce more distinct masks, focusing more on minimizing errors in in-focus regions during later iterations and simultaneously enhancing the quality of the All-in-Focus renderings.
The overall optimization objective is now formulated as:
\vspace{-2mm}
\begin{equation}\label{eq:obj_full}
    \min_{\{\mathcal{S},\:\Omega, \:\{f_m,\: Q_m\}_{m=1}^{M}\}}\mathcal{L}_{\text{rec}} + \mathcal{L}_{\text{detail}} + \lambda_{\text{mk}}\mathcal{L}_{\text{mk}} + \lambda_{\text{reg}}\mathcal{L}_{\text{reg}},
\end{equation}
where $\Omega$ represents the parameters of the ILN, and $\lambda_\text{mk}$ and $\lambda_\text{reg}$ are weights for the loss terms $\mathcal{L}_\text{mk}$ and $\mathcal{L}_\text{reg}$, respectively. 





\section{Experiments and Analysis} 
This section covers our implementation (\Cref{sec:implement}), 
examination of the role of the camera model in our framework (\Cref{sec:analysis}),
results for novel view synthesis under the All-in-Focus setting (\Cref{sec:nvs}), 
and ablation study results (\Cref{sec:ans:ablation}).
 


\begin{figure}[t!]
\begin{center}
\includegraphics[width=0.95\linewidth]{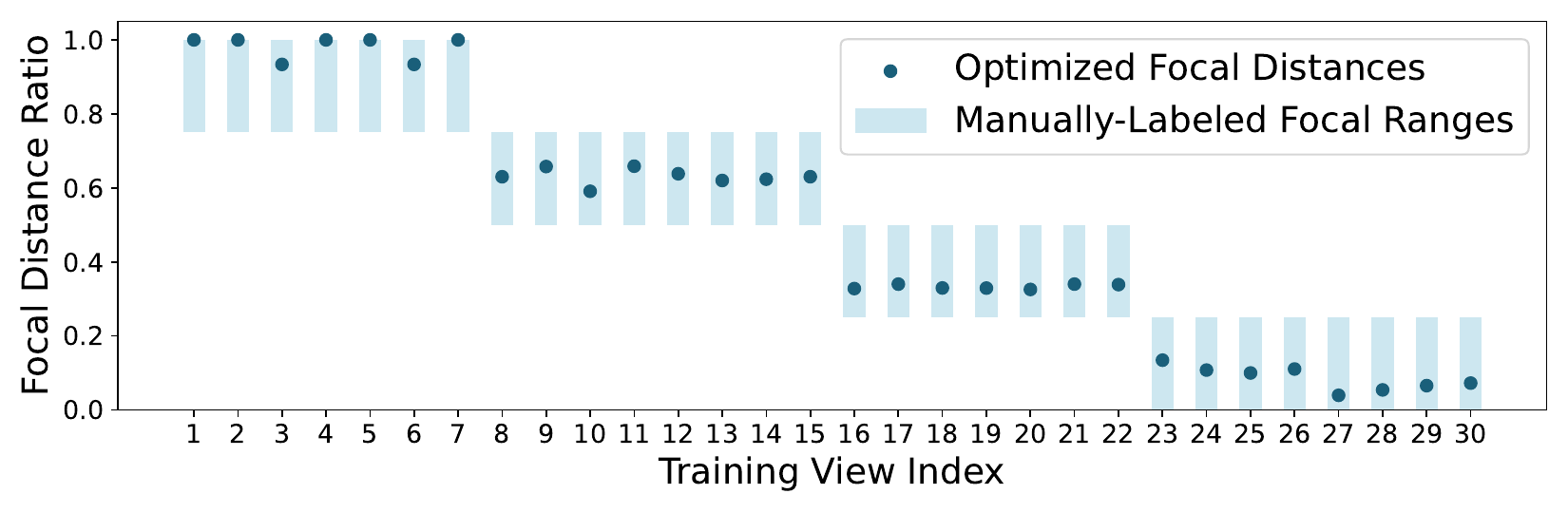}
\end{center}
\vspace{-4mm}
\caption{
\camr{The focal distances optimized using our method closely align with the ranges labeled manually. Focal distances are converted to diopters and normalized for visualization.}
}
\label{fig:focus_vis}
\vspace{-4mm}
\end{figure}
\subsection{Implementation}\label{sec:implement}
The proposed method is implemented using a CUDA-customized rasterizer based on 3DGS~\cite{3dgs} and the PyTorch library \cite{pytorch}. We optimize each scene for $40,000$ iterations. The balancing weights for loss terms are set to: $\lambda_\text{dssim}=0.2$, $\lambda_\text{mk}=0.001$, $\lambda_\text{reg}=0.0001$. 
Further details on the optimization process, including an algorithmic table, are provided in the Supplementary Material.

We train and evaluate our approach using multi-view defocused images from the real and synthetic datasets provided by Ma et al.\cite{ma2022deblur}. The scenes from the real dataset feature manually captured images, with a detailed scene list available in the Supplementary Material. For the synthetic scenes, to mimic real-world conditions where ground-truth camera poses are unavailable, we use camera poses and sparse point clouds estimated by COLMAP \cite{schonberger2016structure, schonberger2016pixelwise} from the provided multi-view blurry rendered images.

\setlength{\tabcolsep}{1.5pt}
\renewcommand{\arraystretch}{0.6}
\begin{figure*}[htbp!]
\begin{center}
\begin{tabular}{cccccc}
\includegraphics[width=.96\linewidth]{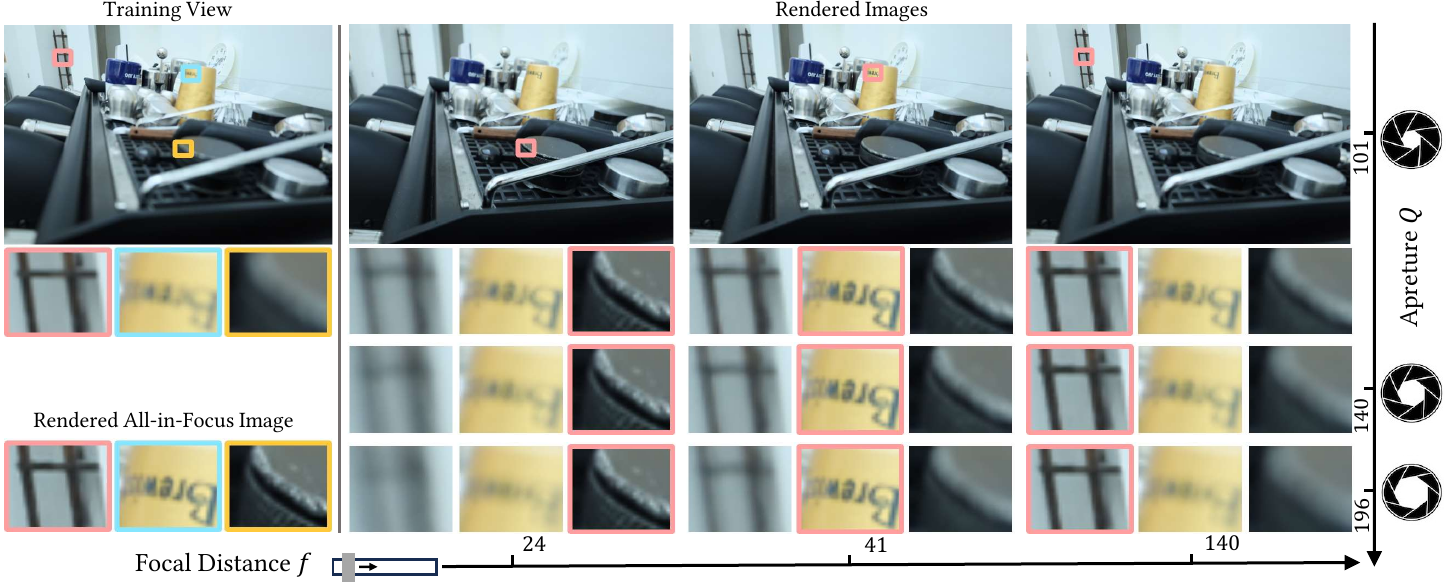}
\end{tabular}
\end{center}
\vspace{-4mm}
\caption{DOF rendering results with post-capture aperture and focal distance control. Adjusting focal distance mainly affects locations of out-of-focus and in-focus regions (highlighted in pink), while increasing aperture parameter makes out-of-focus regions increasingly blurry. 
}
\vspace{-2mm}
\label{fig:dof_render}
\end{figure*}
\subsection{Camera Model Analysis} \label{sec:analysis}
In this section, we analyze the optimized focal distances for each training view, visualize the resulting circle-of-confusion (CoC) maps generated from the optimized camera parameters and the resultant 3D scene, and showcase visual outcomes from independent post-capture adjustments of both the focal distance and aperture size.



\subsubsection{Focal Distance Analysis} 
\label{sec:ans:cam_param}
To validate our method's ability to automatically adapt to training images captured at varying focal distances, we assess the consistency of the optimized focal distances with those in the training views. 
Since precise focal distance labels are unavailable, we sample three scenes and manually annotate a depth range for each view, approximating the camera’s likely focus range during data capture. 
While it is difficult for humans to precisely determine focal distance from a single defocused image, an approximate range can be estimated, especially when comparing multiple images captured at different focal distances.
Consequently, we partition the depth range of objects within each scene into up to four bins, and identify the most likely bin for the focal distance. 
\Cref{fig:focus_vis} shows the optimized focal distances and the annotated ranges, which align closely, thus validating the physical relevance of our camera model. \new{This also validates the effectiveness of our framework in capturing defocus cues without auxiliary annotations.} Additional visualizations can be found in the Supplementary Material.


\vspace{-3.5mm}
\subsubsection{Effect of CoC Map and In-Focus Mask} \label{sec:ans:coc}
\Cref{fig:coc_mask} presents rendered CoC maps, which exhibit a strong correlation with depth-related pixel blurriness, where higher values indicate greater blur. 
This further supports the physical relevance of our defocus rendering process and introduced camera model. \Cref{fig:coc_mask} also displays predicted masks that identify in-focus regions, which are crucial for improving All-in-Focus rendering details. As shown, these masks effectively locate the in-focus areas within training views.




\setlength{\tabcolsep}{0.5pt}
\renewcommand{\arraystretch}{0.6}
\begin{figure*}[t]
\begin{center}
\begin{tabular}{cccccccccc}
\includegraphics[height=2.8cm]{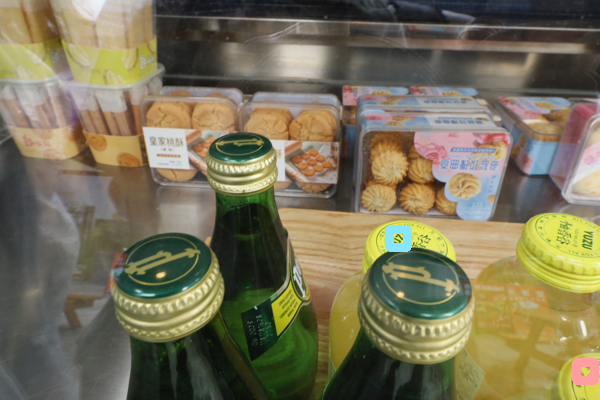}
&\includegraphics[height=2.8cm]{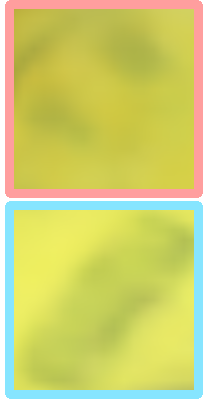}
&\includegraphics[height=2.8cm]{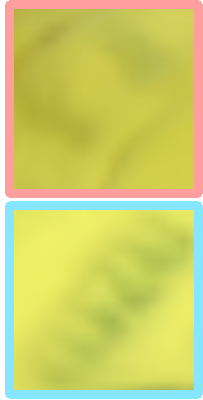}
&\includegraphics[height=2.8cm]{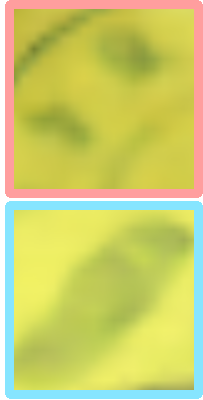}
&\includegraphics[height=2.8cm]{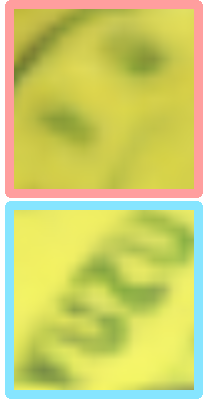}
&\includegraphics[height=2.8cm]{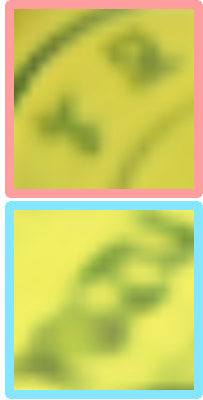}
&\includegraphics[height=2.8cm]{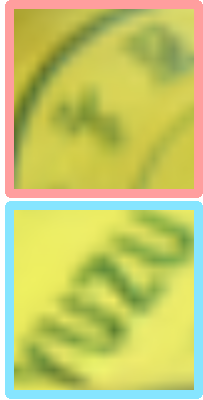}
&\includegraphics[height=2.8cm]{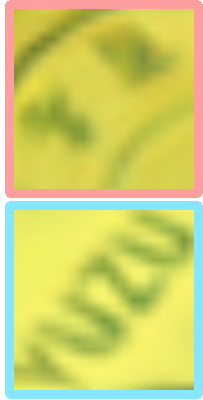}
&\includegraphics[height=2.8cm]{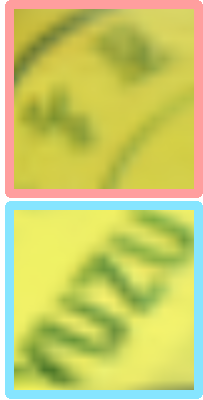}
&\includegraphics[height=2.8cm]{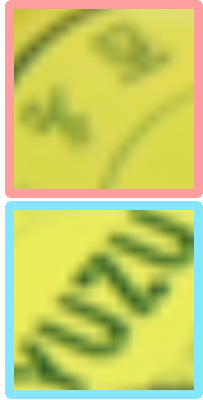}
\\
\includegraphics[height=2.8cm]{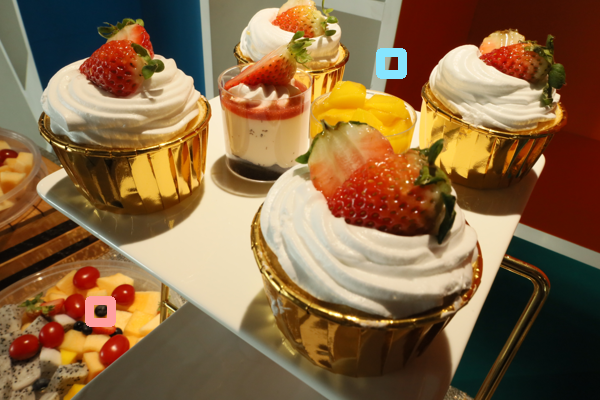}
&\includegraphics[height=2.8cm]{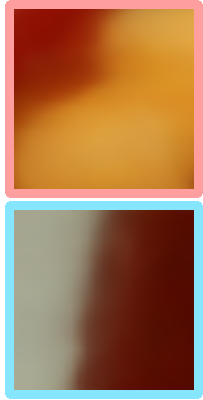}
&\includegraphics[height=2.8cm]{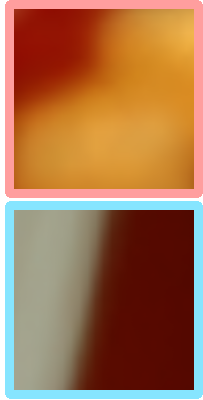}
&\includegraphics[height=2.8cm]{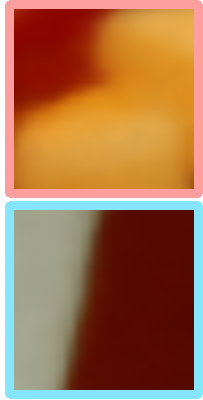}
&\includegraphics[height=2.8cm]{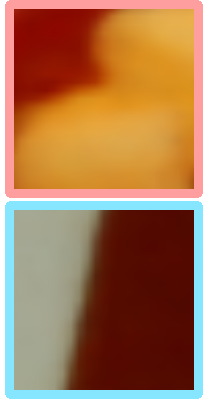}
&\includegraphics[height=2.8cm]{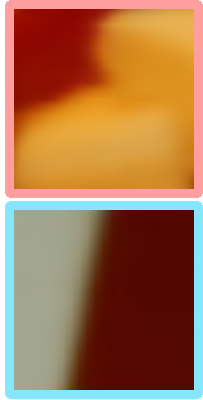}
&\includegraphics[height=2.8cm]{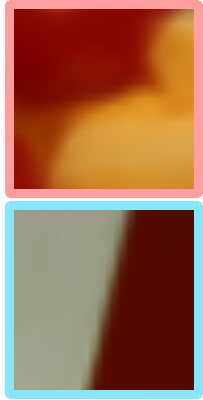}
&\includegraphics[height=2.8cm]{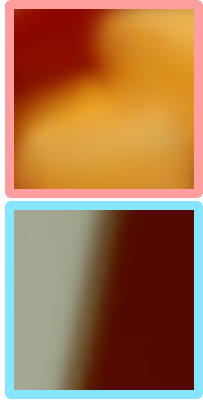}
&\includegraphics[height=2.8cm]{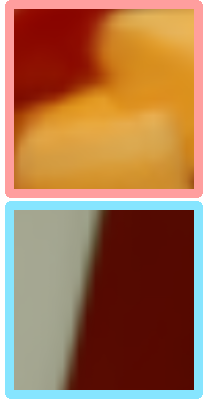}
&\includegraphics[height=2.8cm]{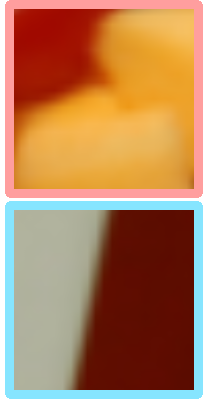}
\\
\footnotesize{Novel View (Reference)} & \footnotesize{NeRF} & \begin{tabular}{c} \footnotesize{Mip-} \\   \footnotesize{Splatting}\end{tabular} & \begin{tabular}{c} \footnotesize{Deblur-} \\   \footnotesize{NeRF}\end{tabular}  & \footnotesize{DP-NeRF} & \footnotesize{PDRF} & \footnotesize{Deblur-GS} & \footnotesize{BAGS} & \footnotesize{Ours}  & \footnotesize{GT} \\  
 \vspace{-7mm}
\end{tabular}
\end{center}
\caption{Visual results for All-in-Focus novel view synthesis. Enlarged regions show our method's ability to recover sharp details.
}
\vspace{-2mm}
\label{fig:visual_comp}
\end{figure*}
\vspace{-2mm}
\subsubsection{Aperture and Focus Control} \label{sec:app}
As discussed earlier, the introduced camera model and defocus rendering process are physically grounded. By effectively learning from the guidance of defocused images, we can simulate a wide range of depth-of-field effects through post-capture adjustments of focal distance and aperture settings. Figures \ref{fig:teaser} and \ref{fig:dof_render} illustrate examples of these effects.
As shown, with a fixed focal distance, the in-focus regions, such as the text region at $f=41$ in \Cref{fig:dof_render}, remain sharp, while out-of-focus regions exhibit varying blur.
When the aperture is kept constant and the focal distance is varied, the locations of in-focus and out-of-focus regions shift accordingly. 
Overall, the blur across different pixels transitions smoothly across images with different focal adjustments, for example, see images for aperture parameter $Q=140$. 
Additional examples can be found in the Supplementary Material. To further demonstrate that our framework supports continuous control over focal distance and aperture---critical for practical applications---we provide video results in the Supplementary Video. These video results show depth-of-field effects rendered from multiple scenes, showcasing continuous adjustments to aperture, focal distance, or both.


\subsection{All-in-Focus Novel View Synthesis} 
\label{sec:nvs}
In addition to enabling post-capture depth-of-field rendering, our framework can also generate sharp All-in-Focus images from reconstructed scenes. We evaluate its performance in novel view synthesis under the All-in-Focus setting ($Q^{*}=0$) and compare it with baseline methods, including
NeRF\cite{mildenhall2020nerf}, Mip-Splatting \cite {mip-splatting}, Deblur-NeRF \cite{ma2022deblur}, DP-NeRF \cite{lee2023dp}, PDRF \cite{peng2023pdrf}, Deblur-GS \cite{lee2024deblurring} and BAGS \cite{peng2024bags}. 
We used the released code for baseline methods for consistency.

\setlength\tabcolsep{1.7pt}
\begin{table}[h]
    \centering
    \scriptsize
    \renewcommand{\arraystretch}{1.1}
    \caption{Numerical results on real and synthetic defocus dataset. We color each cell as \colorbox{red!30}{best}  , \colorbox{orange!30}{second best} and \colorbox{yellow!30}{third best}.}
    \label{tab:average}
    \begin{tabular}{l|c|c|c|c|c|c|c|c}
    \hline
    & NeRF & \begin{tabular}{c}
         Mip-  \\
         Splatting 
    \end{tabular} & \begin{tabular}{c}
         Deblur-  \\
         NeRF 
    \end{tabular} & DP-NeRF & PDRF & BAGS & Deblur-GS & Ours \\ 
    \hline
    \multicolumn{9}{c}{Real Dataset (Res.: $600 \times 400$)} \\
    \hline
    PSNR & 22.70 & 21.23 & 23.98 &  24.12 & \cellcolor{red!30} 24.31 & \cellcolor{yellow!30} 24.17 & \cellcolor{orange!30} 24.21 & 24.12 \\ 
    SSIM  & .6789 & .6292 & .7388 & .7451 &  .7559 & \cellcolor{orange!30}.7664& \cellcolor{yellow!30} .7631 & \cellcolor{red!30} .7667 \\ 
    \hline
    \multicolumn{9}{c}{Synthetic Dataset (Res.: $600 \times 400$)} \\
    \hline
    PSNR  & 25.74 & 23.73 & 28.02 & 28.91 & \cellcolor{red!30} 29.96 & 28.89 & \cellcolor{orange!30} 29.43 & \cellcolor{yellow!30} 29.39 \\ 
    SSIM  & .7769 & .7314 & .8491 & .8687 & \cellcolor{red!30} .8923 & .8859 & \cellcolor{yellow!30} .8907 & \cellcolor{orange!30} .8914 \\ 
    \hline
    \end{tabular}
\end{table}

\camr{We report the average performance on real scenes and synthetic scenes in \Cref{tab:average}. Our method performs comparably to existing deblurring NeRF and 3DGS methods. The comparable performance may result from the stricter constraints of the two-parameter camera model's global DOF modeling, which reduces flexibility compared to per-pixel kernel estimation.}
Sampled visual results are provided in \Cref{fig:visual_comp}. As shown, Mip-Splatting, like vanilla NeRF, struggles to reconstruct fine details due to significant defocus blur in the input images. 
By contrast, our approach consistently outperforms or matches the performance of state-of-the-art NeRF-based and 3DGS-based methods, 
recovering details in challenging areas such as 
text, fruit stacks, and wall patterns.
This improvement is due to our DOF rendering, which maintains differentiablity and effectively models defocus blur, thereby reducing its impact on scene reconstruction.
Comprehensive quantitative comparisons and additional visual results are available in the Supplementary Material.

\setlength\tabcolsep{0.8pt}
\renewcommand{\arraystretch}{1.2} %
\begin{table}[t!]
\centering
\scriptsize
\vspace{-1.6mm}
\caption{Numerical results on real defocus dataset at \textbf{higher} resolutions. We color each cell as \colorbox{\bestcolor}{best} and \colorbox{\secondbestcolor}{second best}}
\begin{tabular}{l|c|c|c|c|c|c}
\cline{1-7}
\multirow{2}{*}{Method} & \multicolumn{3}{c|}{Res.: $1200\times800$} & \multicolumn{3}{c}{Res.: $2400 \times1600$} \\
\cline{2-4} \cline{4-7}       
& \makecell{ Avg. \\ PSNR $\uparrow$}  & \makecell{ Avg. \\ SSIM $\uparrow$} &  \makecell{ GPU Mem. \\ Usage (GB) $\downarrow$}  & \makecell{ Avg. \\ PSNR $\uparrow$}  & \makecell{ Avg. \\ SSIM $\uparrow$}  & \makecell{ GPU Mem. \\ Usage (GB) $\downarrow$}  \\
\cline{1-7} 
Mip-Splatting & 20.51  & .5995 & \cellcolor{\bestcolor}8.75 & 20.35 & .6220  & \cellcolor{\bestcolor}32.13\\ 
BAGS & \cellcolor{\secondbestcolor}23.60 & \cellcolor{\secondbestcolor}.7345 & 21.69 &  - & -   & - \\
\cline{1-7}
w/o. detail enhance  & 23.36 & .7306 & \cellcolor{\secondbestcolor}9.56 & \cellcolor{\secondbestcolor}23.07 & \cellcolor{\secondbestcolor}.7190 &  \cellcolor{\secondbestcolor}38.89  \\
Full Model  & \cellcolor{\bestcolor}23.66 & \cellcolor{\bestcolor}.7376 & 13.06   & \cellcolor{\bestcolor}23.45 & \cellcolor{\bestcolor}.7221 &  39.05\\
\cline{1-7}
\end{tabular}%
 \begin{tablenotes}
\item{ \footnotesize{-: Results are unavailable due to excessive memory required for training.}}
\item{ \footnotesize{The 3rd and 4th rows show results from two variants of our approach.}}
\end{tablenotes} 
\vspace{-3mm}
\label{tab:real_high_res}%
\end{table}%

\paragraph{Performance with Increasing Resolution}
\label{sec:nvs:larger}
To evaluate our method’s adaptability to high-resolution inputs, we test it at resolutions of $1200\times800$ and $2400\times1600$. \Cref{tab:real_high_res} summarizes peak GPU memory usage and performance comparisons with two other 3DGS-based methods, Mip-Splatting \cite{mip-splatting} and BAGS \cite{peng2024bags}. Our method performs robustly at both resolutions, with only a modest memory increase compared to Mip-Splatting. 
While the deblurring-focussed BAGS achieves similar quality at $1200\times800$, it demands $166\%$ more memory. The memory efficiency of our approach is attributed to our defocus blur modeling, which mathematically applies depth-based Gaussian blurs directly to Gaussian points. 
Additionally, our variant, \emph{Ours (w/o. detail enhance)}, conserves memory by disabling the All-in-Focus enhancement at a slight performance cost.
These results highlight our method’s strong adaptability to high-resolution inputs.

\subsection{Ablation Study} \label{sec:ans:ablation}
\new{In ablation studies, we evaluate the quantitative performance of All-in-Focus novel view synthesis for several variants.
Results in \Cref{tab:ablation}
indicate that our detail enhancement strategy via imposing supervision on rendered All-in-Focus images
significantly enhances PSNR and SSIM scores, confirming its effectiveness.}
\camr{Supplementary visual examples further illustrate improved details and reduced artifacts. 
The variants \emph{Ours} (w/o. $\mathcal{L}_{\text{mk}}$) and \emph{Ours} (w/o. $\mathcal{L}_{\text{reg}}$) achieve similar PSNR scores to the full model but show declines in SSIM, suggesting the positive impact of regularizing in-focus masks utilizing CoC maps and the binary regularizer $\mathcal{L}_{\text{reg}}$, as evidenced by the visual results in the Supplementary Material. }

\setlength\tabcolsep{3.5pt}
\renewcommand{\arraystretch}{0.97} %
\begin{table}[t!]
\centering
\footnotesize
\vspace{-1.6mm}
\caption{Ablation of DOF-GS on real defocus dataset.}
\vspace{-1mm}
\begin{tabular}{l|c|c|c|c|c}
\cline{1-6}
Variant & \multicolumn{3}{c|}{Ablation Items} & \multicolumn{2}{c}{Average Scores} \\
 \cline{2-6}
 & $\mathcal{L}_{\text{detail}}$ & $\mathcal{L}_{\text{mk}}$ & $\mathcal{L}_{\text{reg}}$ & PSNR $\uparrow$ & SSIM $\uparrow$ \\
 \cline{1-6}
Ours (w/o. detail enhance) & \ding{55} & \ding{55} & \ding{55}  & 23.36 & 0.7306 \\
Ours (w/o. $\mathcal{L}_{\text{mk}}$) & \ding{51} & \ding{55} & \ding{51}  & 24.09 & 0.7642 \\
Ours (w/o. $\mathcal{L}_\text{reg}$) & \ding{51} & \ding{51} & \ding{55}  & 24.09 & 0.7640 \\
\cline{1-6}
Ours (Full Model) & \ding{51} & \ding{51} & \ding{51} & \textbf{24.12} & \textbf{0.7667} \\
\cline{1-6}
\end{tabular}%
\vspace{-3mm}
\label{tab:ablation}%
\end{table}%

\section{Conclusion}

We present DOF-GS, a 3D Gaussian Splatting framework that integrates a finite-aperture camera model to 
enable refocusing and depth-of-field effects at runtime.
DOF-GS effectively reconstructs high-quality 3D scenes from uncalibrated, defocused multi-view inputs, and achieves superior all-in-focus novel view synthesis with minimal increase in GPU memory.
In particular, our method supports post-capture adjustments to camera parameters such as aperture and focal distance, enhancing imaging capabilities. \new{Recovering sharp details for regions that remain blurry in all views remains challenging.
A future direction is to incorporate camera pose optimization during training, as this could improve camera pose accuracy and, consequently, scene reconstruction quality in the presence of defocus blur.}
We anticipate our work to spark further advances in flexible, high-fidelity and high-dimensional imaging, scene modeling and analysis.

\section*{Acknowledgment}
This work was partially supported by the National Key R\&D Program of China (Grant No. 2022ZD0160801) and NSF Grant 2107454. We sincerely thank Guangyuan Zhao and Yuanxing Duan for fruitful discussions.

{\small
\bibliographystyle{ieee_fullname}
\bibliography{reference}
}

\end{document}